\documentclass{ecai}
\usepackage{times}
\usepackage{graphicx}
\usepackage{latexsym}
\usepackage{amsmath}
\usepackage{amssymb}
\usepackage{subcaption}
\usepackage{gensymb}

\DeclareMathOperator*{\argmax}{arg\,max}

\begin{document}

\title{Modeling Rare Interactions in Time Series Data Through Qualitative Change: Application to Outcome Prediction in Intensive Care Units} 

\author{Zina Ibrahim \institute{King's College London, email: zina.ibrahim@kcl.ac.uk }
\and Honghan Wu \institute{University of Edinburgh, email: honghan.wu@ed.ac.uk} \and Richard Dobson \institute{King's College London, email: richard.j.dobson@kcl.ac.uk }
 }

\maketitle
\bibliographystyle{ecai}

\begin{abstract}
Many areas of research are characterised by the deluge of large-scale highly-dimensional time-series data. However, using the data available for prediction and decision making is hampered by the current lag in our ability to uncover and quantify true interactions that explain the outcomes. 
We are interested in areas such as intensive care medicine, which are characterised by i) continuous monitoring of multivariate variables and non-uniform sampling of data streams, ii) the outcomes are generally governed by interactions between a small set of rare events, iii) these interactions are not necessarily definable by specific values (or value ranges) of a given group of variables, but rather, by the deviations of these values from the normal state recorded over time,  iv) the need to explain the predictions made by the model. Here, while numerous data mining models have been formulated for outcome prediction, they are unable to explain their predictions.

We present a model for uncovering interactions with the highest likelihood of generating the outcomes seen from highly-dimensional time series data.  Interactions among variables are represented by a relational graph structure, which relies on qualitative abstractions to overcome non-uniform sampling and to capture the semantics of the interactions corresponding to the changes and deviations from normality of variables of interest over time. Using the assumption that similar templates of small interactions are responsible for the outcomes (as prevalent in the medical domains), we reformulate the discovery task to retrieve the most-likely templates from the data. Experiments on sepsis prediction using real Intensive Care Unit (ICU) data demonstrates that the discovered interaction templates are semantically meaningful within the domain, and using them as features in a prediction task produces a superior performance than when using the raw values of the predictors. 

\end{abstract}

\section{Introduction} \label{sec:introduction}
Many of the big-data sources now available correspond to complex entities embedding sophisticated interactions recorded over time (e.g. a potential diagnosis is affected by multiple symptoms, changes in a patient's vital signs and response to ongoing treatments, among others). Such data sources are characterised by highly granular observations (e.g. ICU patients are monitored as frequently as every 5 seconds), and uneven reporting of data points (e.g. in the ICU, respiration rate is measured hourly, while heart rate is automatically recorded every 5 seconds). Nevertheless, a small number of interactions (or rare events) may be the most important ones for predicting outcomes and explaining deviations from usual functioning (e.g. rare events in the ICU signal critical changes in a patient's state). Discovering such interactions is especially crucial in areas where the ultimate goal is to monitor entities in real-time, to alert users when abnormal behaviour of the system is present  (e.g. a potential adverse event for a patient). 

This work starts with the observation that interactions among a set of temporal variables are usually accompanied by a joint change in these variables' qualitative states (e.g. increase, low). Although the metric information accompanying interactions is often unique (e.g. specific values of the heart rate when body temperature exceeded 37.5$^0C$), abstracting away from the numerical details of event occurrence renders such interactions comparable. Abstraction aggregates interactions into meaningful, qualitative concepts that can hold over time intervals, by describing the changing effect of a given interaction as well as the associated temporal arrangements. Such thinking is consistent with the qualitative way of reasoning adopted in the medical domain \cite{medicalqualitative1,medicalqualitative2,medicalqualitative3,medicalqualitative4,medicalqualitative5}. For example, in addressing the proper treatment of patients, including effects of drugs and volume input, time is expressed in the relational model between time intervals, abstracting away the exact time of occurrence and duration (e.g. a period of high and increasing heart rate accompanied by periods of low and decreasing respiratory rate). Moreover, such representation is free from the following unrealistic assumptions made by many temporal data mining models: (1) perfectly aligned atomic patterns can be obtained, and (2) patterns to be discovered are of equal lengths  \cite{hu2013}. In reality, these two assumptions cannot be made when analysing medical data. For example, when analysing arrhythmia episodes in 10-minute ECG scans, such events rarely occur at the same time point in all patients, nor will they have identical durations.

\begin{figure*}[!ht]
    \centering
    \begin{subfigure}[t]{0.7\textwidth}
        \centering
        \includegraphics[width=\textwidth]{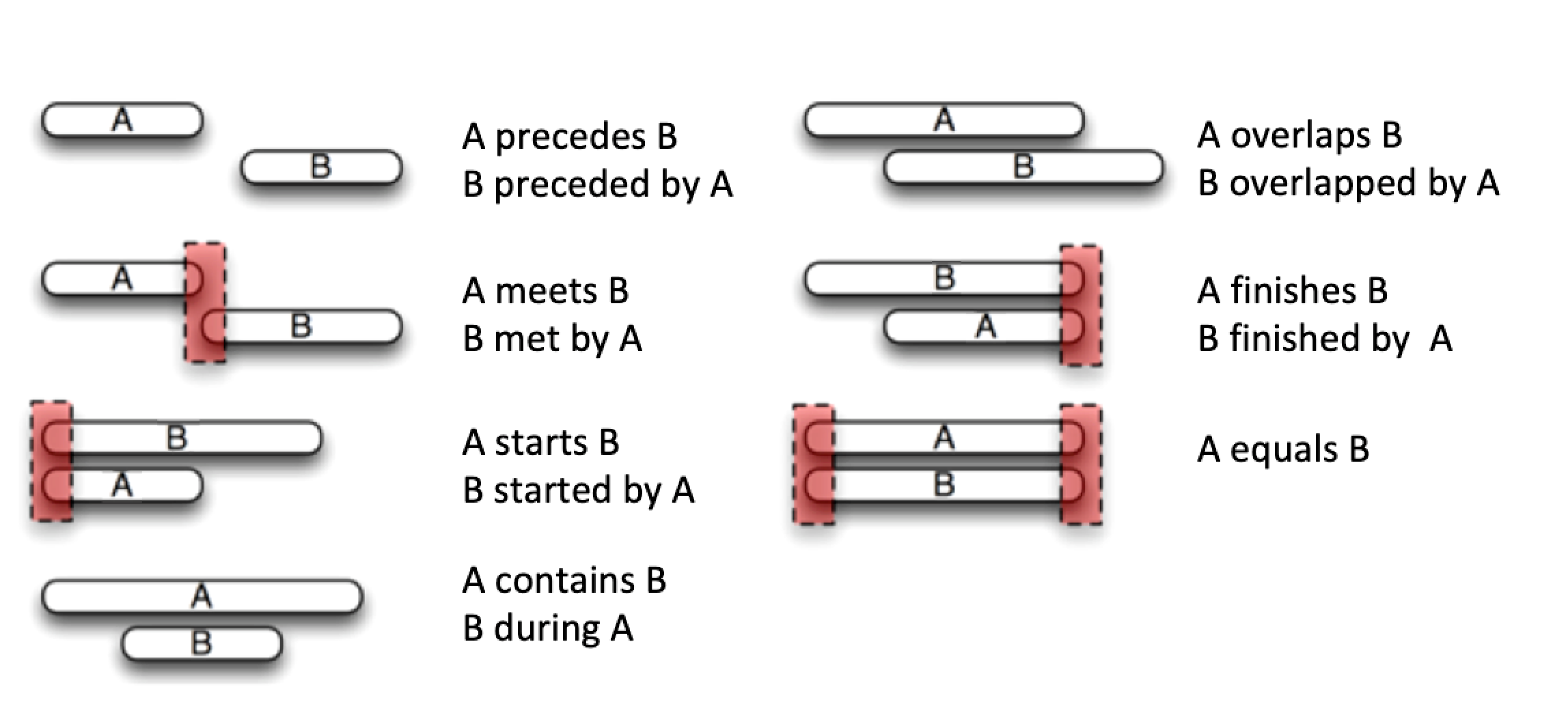}
        \caption{}
    \end{subfigure}%
    ~ 
    \begin{subfigure}[t]{0.3\textwidth}
        \centering
        \includegraphics[width=.7\textwidth]{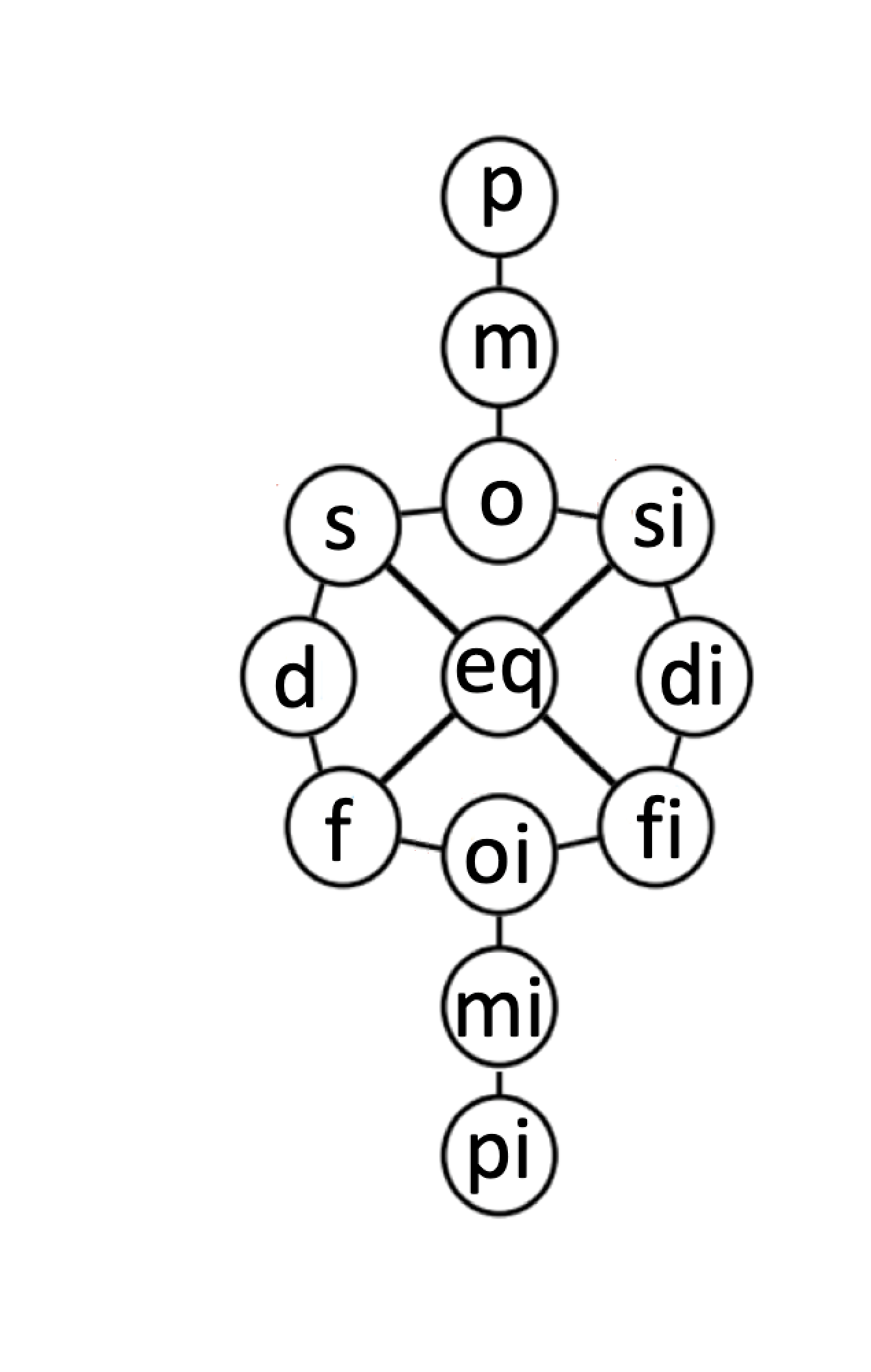}
        \caption{}
    \end{subfigure}
    
    \caption{\textbf{(a)} Allen's base seven temporal relations and their inverses. \textbf{(b)} The conceptual neighbourhood created by enforcing temporal continuity over Allen's 14 relations. The conceptual neighbourhood graph shows the permitted transitions from one relation to another while preserving temporal continuity. In the figure: p: precedes; m: meets; o: overlaps; s: starts,;d: during; f: finishes; pi: proceded by; si: started by; mi: met by; oi: overlapped by;  di: contains; fi: finished by; eq: equals. }
\label{fig:allen} 
\end{figure*}

In this paper, we present and evaluate a framework for discovering interaction patterns in time series data by capturing \emph{qualitative signatures} embedded within the data, subsequently using those for outcome prediction. We formulate the concept of a \emph{qualitative interaction graph}, in which nodes correspond to qualitative change descriptions of events and edges represent interactions, by encoding qualitative temporal relations that hold between events. As most resulting interactions in a qualitative interaction graph may indicate normal functioning or are merely coincidental, our aim is to uncover the \emph{significant interactions} that play an integral part in generating the outcomes. We do this by defining a generative model for significant interactions upon which we perform an expectation-maximisation procedure to iteratively evaluate the significance of found interactions with respect to a given outcome.

The importance of the framework lies in its ability to extract intuitive and multi-variable patterns of deviation from normal behaviour, capturing rare events in highly-dimensional, multivariate and non-uniformly sampled data. The framework retains all the useful information required for outcome prediction without jeopardising performance. Moreover, and as opposed to the state of the art approaches, the framework can explain the outcomes predicted through the use of the aforementioned qualitative patterns.

This paper is structured as follows. After discussing related work in Section \ref{sec:related}, we delve into the knowledge representation aspect of our work in Sections \ref{sec:qig} and \ref{sec:describe}. We then formulate a pattern-discovery model based on the Expectation-maximisation (EM) algorithm in Section \ref{sec:generate}. In Section \ref{sec:experiments}, we evaluate the performance of the pattern discovery model using sepsis as a case study and real ICU data. We perform two sets of experiments: while the first experiment focuses on the discovery of significant interactions in a given population, the second is a classification experiment which verifies the ability of discovered interactions to discriminate patients who will develop sepsis from those who will not, eight hours before onset. Finally, we discuss the status of work and ongoing efforts in Section \ref{sec:conclusion}. 

\section { Related Work}\label{sec:related}

The literature contains several data mining models that represent time-stamped multivariate raw data as a set of time intervals, often at a higher level of abstraction, and subsequently use discovered temporal patterns for classification tasks \cite{moskovitch2015,patel2008,shahar2017,shahar2015,aime,ibrahim}. 

Most of these interval discovery models use Allen's seminal work \cite{allen1983}, which devises binary relations to capture the order and interactions between temporal intervals.  Allen's interval algebra contains a total of thirteen mutually-exhaustive and pairwise-disjoint qualitative relations, by which the temporal relationship between any two events can be unambiguously described. These relations are given in Figure \ref{fig:allen} (a). The set consists of six basic relations: precedes, meets, starts, contains, overlaps, finishes, and their inverses: preceded by, met by, started by, during, overlapped by and finished by. In the case of the equal relation, the basic and inverse relations are identical. The notion of an algebra over these relations arises from considering the intersection, union, and composition of a pair of temporal relations \cite{allenalgebra}.  Allen's algebra enforces the notion of \emph{temporal continuity} via the relations' \emph{conceptual neighbourhood} \cite{freska}. In a conceptual neighbourhood, two relations between pairs of events are conceptual neighbours if they can be directly transformed into one another by continuous deformation (i.e., shortening or lengthening) of the events. Allen's conceptual neighbourhood structure is thus obtained and is shown in Figure  \ref{fig:allen} (b). In the figure, the intervals are replaced by circles containing the symbolic abbreviations of the names of the corresponding relations as given in Figure  \ref{fig:allen} (a). Solid lines depict neighbourhood relations. The \emph{conceptual neighbourhood distance} between two qualitative relations quantifies the notion of temporal continuity. It is defined as the shortest path between these relations in the conceptual neighbourhood graph, giving every arc a distance being equal to one. Given two relations $r_i$ and $r_j$, $d(r_i, r_j)$ is the conceptual neighbourhood distance between the two relations and has a minimum value of zero (when the two relations are the same). 

In addition to using Allen's algebra, most of the existing data mining models use Shahar's knowledge-based temporal abstraction framework \cite{shaharkb}, which captures, among other properties, the qualitative states (e.g. high, low, normal) and qualitative gradient changes (e.g. increasing, decreasing, constant) during temporal intervals.

However, existing models that use qualitative temporal abstractions to mine time-series data, exemplified by \cite{patel2008,shahar2015}, have two vital shortcomings. First, they individually mine qualitative state (e.g. high) and gradient (e.g. increasing) changes, without capturing the richer semantics resulting from the simultaneous representation of state and gradient changes as temporal patterns, and examining the temporal interactions generated. This is especially vital for the medical domain (and the ICU in specific) where events are described by multiple simultaneous qualitative descriptors. Second, they generally focus on mining frequent temporal patterns, with the assumption that they are the highest contributors to a given outcome; an assumption that does not hold in the medical domain as we illustrated in Section \ref{sec:introduction}. Nevertheless, we build on the knowledge generated by these approaches, using the knowledge-based temporal abstraction model of \cite{shaharkb} as a foundation for defining the qualitative concepts used in our work. In addition, we borrow the evaluation approach presented in \cite{moskovitch2015} to show that using our discovered interactions can be used as features in classification problems to achieve high performance. Finally, we borrow the variable abstraction knowledge base of \cite{shaharsepsis} in the processing of our data in Section \ref{sec:experiments}. 

The literature also contains models which aim at the discovery of causal relations underlying a given large data \cite{pearl,murphy2002,kleinberg2009,kleinberg2011,kleinberg2012}. However, many of the causality work requires the specification of prior probabilities of events, which is not possible for many domains  - for example, in the ICU. To overcome the inference difficulties, causal inference models using the expected values of domain variables to detect rare events have been formulated \cite{causal}, but are yet to take into account the specific knowledge representation requirements of intensive care. However, we acknowledge the importance and impact of \cite{causal}, and adapting its inference model to accommodate our knowledge representation framework is part of our ongoing work.

\section{Qualitative Interaction Graphs}\label{sec:qig}

The basic idea of the approach is to generate a set of patterns of qualitative change embedded within time-series data and represent their potential interactions via a graphical representation termed a Qualitative Interaction Graph (QIG). 


\subsection{ Knowledge-based Temporal Abstraction (KBTA)} 
\begin{figure}[!ht]
\centerline{\includegraphics[width=0.5\textwidth]{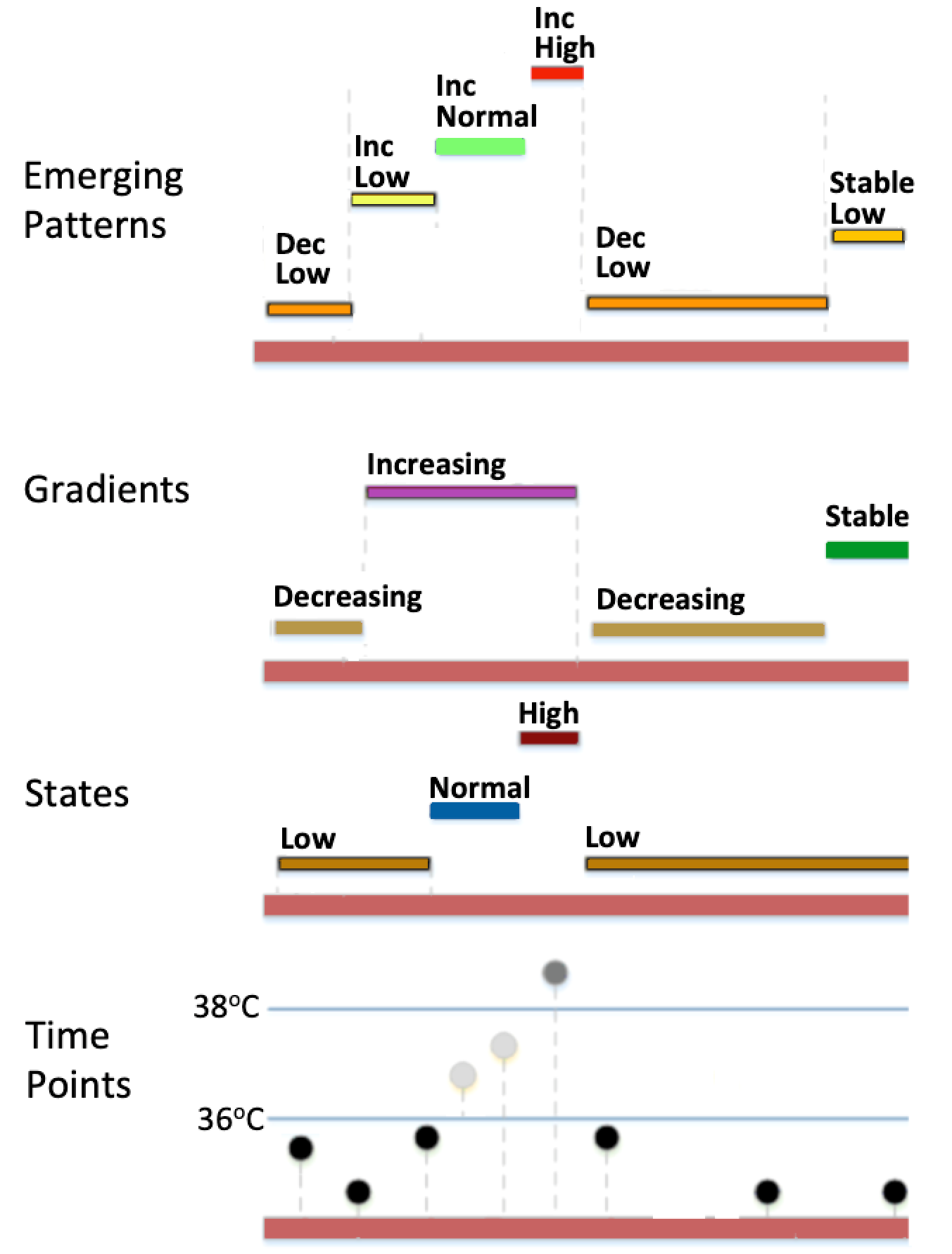}}
\caption{
A series of raw time-point data for a patient's body temperature is presented at the bottom. The data in this case are abstracted according to their values into four interval-based states (layer 2 from the bottom), and into four gradient abstractions (layer 3 from the bottom). The resulting temporal patterns combining state and gradient abstractions are shown in the top layer of the figure. There are exactly six temporal patterns resulting from the joint abstractions of state and gradient changes. } 
\label{fig:abstraction}
\end{figure}

Given time-stamped raw data grouped by objects of interest (e.g. a patient's ICU stay) and comrpising a set of temporal domain variables $V$ (e.g. all vital signs recorded in an ICU over a period of time), a set of abstract interval-based concepts are obtained for the object of interest by exploiting domain-specific knowledge. We use the knowledge-based temporal abstraction (KBTA) model introduced by Shahar \cite{shaharkb}. The model uses cut-off values suggested in a context-sensitive fashion by a domain expert to determine maximal intervals whereby qualitative state changes (low, normal, high, e.g. high heart rate) as well as qualitative gradient changes (increasing, decreasing, stable, e.g. decreasing respiratory rate) hold. The bottom three layers of Figure \ref{fig:abstraction} show an example of KBTA for body temperature. The numerical values are given in the lowest level of the figure,  while the second and third levels show the state and gradient abstractions respectively.

\subsection{ Pattern Creation}  Maximal intervals of qualitative state-gradient pairs are aggregated into a sequence of temporally-contiguous patterns, such that within each pattern, the same state and gradient relations hold during the pattern's interval, but not immediately before or after the interval. The state-gradient temporal patterns abstracting body temperature are shown in the top layer of Figure \ref{fig:abstraction}. In the figure, the interval during which the gradient abstraction \emph{Increasing} holds results in three qualitative state-gradient patterns, by combining the gradient abstraction with the state abstractions that hold during the same interval. The resulting patterns are: \emph{Increasing-Low}, \emph{Increasing-Normal} and \emph{Increasing-High} shown in the top layer of the figure.

\subsection{Pattern Template Creation} For each single variable (e.g. body temperature) $V_z \in V$ where $V$ is the set of temporal domain variables for an object of interest (e.g. all vital signs recorded in an ICU over a period of time for a given patient), we distinguish a qualitative pattern from a qualitative pattern template. Patterns are aggregated into pattern templates such that two patterns map to the same template if they have the same qualitative state and gradient descriptions. Therefore, for a single variable, the corresponding set of pattern templates comprises unique qualitative state-gradient descriptors. For now, the start and end time of a template is the enumeration of the start and end times of the patterns making up the template. In Figure \ref{fig:abstraction}, two patterns of \emph{Decreasing-Low} map to the same pattern template, resulting exactly five pattern templates in the figure: \emph{Decreasing-Low}, \emph{Increasing-Low}, \emph{Increasing-Normal}, \emph{Increasing-High},  and \emph{Stable-Low}. 

\subsection{Generating Qualitative Interaction Graphs } 
\begin{figure}[!ht]
\centerline{\includegraphics[width=0.5\textwidth]{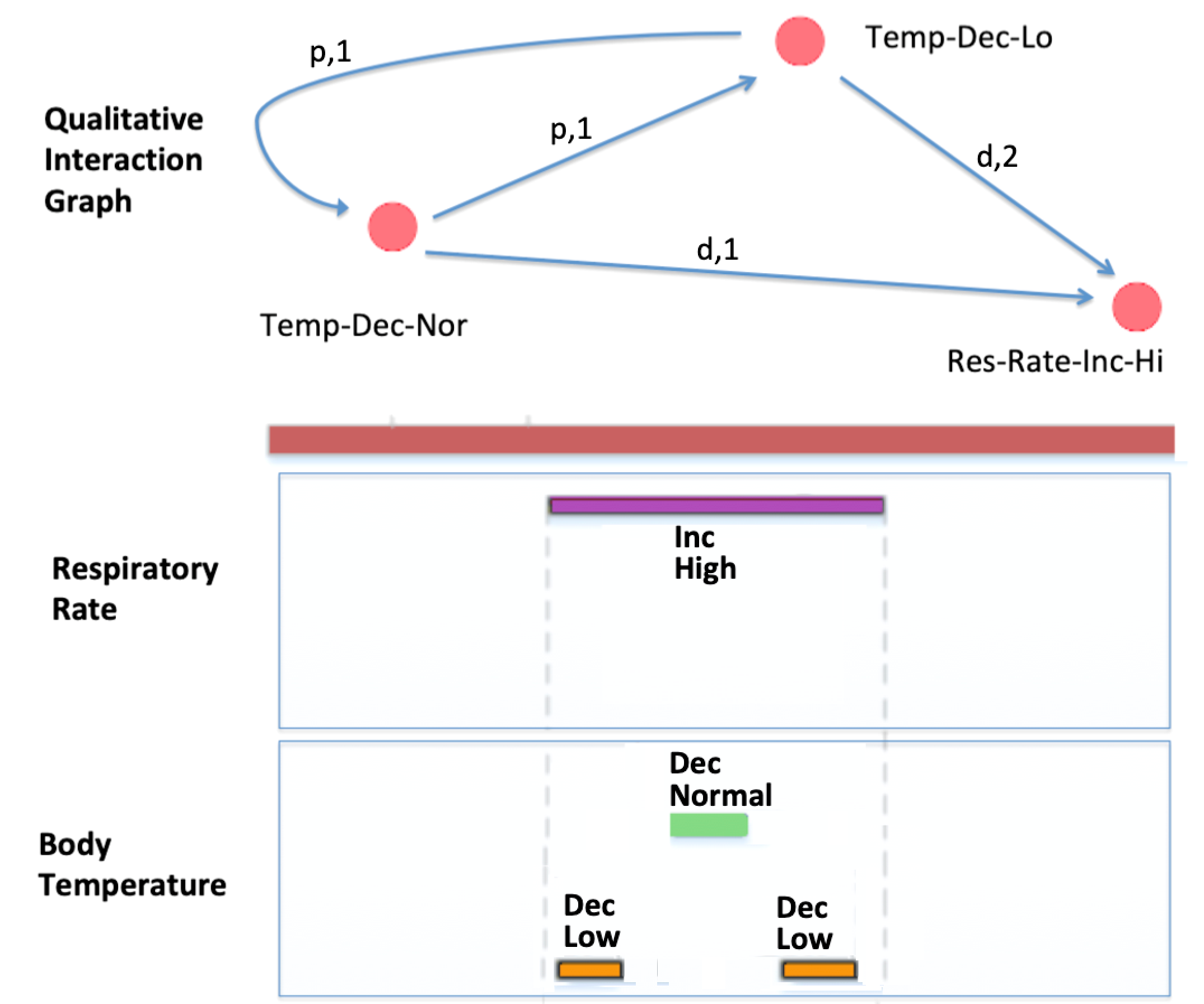}}
\caption{\label{fig:graph}
A Qualitative Interaction Graph connecting the pattern templates of respiratory rate and body temperature. There are two patterns which map to the same pattern template: \emph{Temp-Dec-Low}; both occur during the \emph{Res-Rate-Inc-Hi} template, rendering the connecting edge weight 2.} 
\end{figure}

Given a set of temporal domain variables $V$, such that the state-gradient abstractions of each variable $V_z \in V$ correspond to a set pattern templates $P_z$, qualitative temporal relations between pattern templates of different variables gives rise to a qualitative interaction graph $G=(P, R)$. Figure \ref{fig:graph} shows the creation of a QIG (top of the figure) from pattern templates of body temperature and respiration rate (bottom of the figure). A QIG $G$ has the following properties: 

\begin{enumerate}
    \item $G$ is a connected node and edge-labelled multi-graph in which the nodes $P$ in which each node correspond to all pattern templates of all temporal variables of interest extracted from the data timeline of a given object, and the edges $R$ correspond to the qualitative temporal relations holding between any two pattern templates. 
    \item  Interactions among variables are captured by the temporal relationships between their corresponding pattern templates. The edges reflect the semantics of the corresponding interaction via: 
    \begin{enumerate}
        \item Edge labels, which describe the qualitative temporal relation connecting the interacting pattern templates.
        \item Edge weights, which capture the frequency of a given interaction between two templates. 
    \end{enumerate}
     \item $G$ a multi-graph, with multiple edges between two pattern templates denote different  temporal interactions between the two templates. 
     \item $G$ is a directed graph, preserving the semantics of Allen's interval relations. We use Allen's seven base relations described in Figure \ref{fig:allen}. This means that we do not use inverse relations (as they are implicitly defined by edge reversal), and limit the \emph{before} relation by a maximal allowed gap, as previously done in \cite{shahar2015,winarko}. 

\end{enumerate}

Therefore, the structure of a QIG captures the simultaneous change in state and gradient abstractions of the domain variables for a given object (e.g. patient).

\section{Describing and Comparing Qualitative Interaction Graphs}\label{sec:describe}

We follow an approach where a qualitative interaction graph is described by a collection of subgraphs, where each subgraph captures several possible temporal interactions between subsets of qualitative templates as described in Section \ref{sec:qig}. We begin by populating a subgraph lexicon $li$ of $n$ elements, to contain all possible subgraphs for a given collection of variables, by first enumerating all pattern templates and then generating subgraphs of increasing number of nodes to some fix bound (for a total of $n$ subgraphs in the lexicon). Using the subgraph lexicon, the structure of a qualitative interaction graph of a single object $i$ (e.g. patient) can be represented via a multi-hot vector of length $n$:

\begin{equation*}
\theta(G_i) = [li(k) \subseteq G_i; \forall k \in 1...n ]
\end{equation*}

The multi-hot encoding of a given graph captures the structural properties of the graph as described by the graph lexicon. When evaluating the similarity of the two graphs (e.g. to evaluate the similarity of the treatment journeys of two patients), we not only account for the extent to which they share subgraphs (captured by their corresponding multi-hot vectors), but also take into account the semantics of the interactions, captured by the weights and labels of the subgraph edges. For any two edges, we define their similarity as: 

\begin{equation}
    sim(E_i, E_j)  = E_j \asymp E_j \times \frac{1} {d(r_i, r_j)+ |w_{i}-w_{j}| +1} 
\end{equation}

Where $E_j \asymp E_j$ denotes the correspondence between the nodes connecting each edge, and results in 1 when the nodes connecting the two edges are the same, and zero otherwise. $d(r_i, r_j)$ is the conceptual neighbourhood distance between the temporal relations $r_i$ and $r_j$ (labeling edges $E_i$ and $E_j$ respectively) described in Section \ref{sec:related}. This enables the similarity function $sim$ to take into account varying degrees of similarity, with the maximum value being 1 (when the two edges are labeled with the same temporal relation, making  $d(r_i, r_j)=0$, and have equal weights, making $|w_{i}-w_{j}| = 0$). 

We can therefore define the \emph{similarity vector} of two graphs $G_i$ and $G_j$ representing the normalised similarity of the subgraphs embedded in the two graphs' multi-hot vectors: 

\begin{multline}
    \displaystyle \Theta(G_i, G_j)  =  
    [\frac{\sum_{\substack{E_v \in G_i \\ E_w \in G_j } }  sim(E_v, E_w)}{|R(G_i)|} \\ ,if \theta(G_i)_k = 1 \wedge \theta(G_j)_k = 1;  \forall k \in 1 ...n  ]
\end{multline}

Normalising the similarity of two subgraphs in the above equation, by dividing it by their mutual number of edges $R(G_i)$, guarantees that the similarity value will be in the range $[0,1]$ and is not influenced by the size of the subgraphs at position $k$. 

Finally, the similarity between two graphs $\mathcal{S}(G_i, G_j) $ $\rightarrow [0,1]$ is the normalised sum of the two graphs' similarity vector. 

\begin{equation}
    \mathcal{S}(G_i, G_j) = \displaystyle\frac{\displaystyle\sum_{k=1}^n \Theta(G_i, G_j)}{n}
\end{equation}

\section{A Framework for Uncovering Rare Interactions}\label{sec:generate}

The aim of this work is to establish the grounds for finding interactions that are embedded within a temporal dataset, and have the highest contribution to generating a given outcome. In order to distinguish such significant interactions from all other interactions which may indicate normal functioning or spurious events, the following model is assumed to underlie the generation of significant interactions for the domain of interest. 

\subsection{Prior Distribution Over Pattern Templates}

We describe the underlying prior probability distribution $P(\mathcal{G})$ over sets of potential interaction subgraphs $\boldsymbol{\mathcal{G}} = \{{G}_1, ....{G}_m\}$ as an exponential distribution: 

\begin{equation}
    P(\boldsymbol{\mathcal{G}}) = exp( - \lambda_1 \mathcal{W}(\boldsymbol{\mathcal{G}}) + \lambda_2 \mathcal{N}(\boldsymbol{\mathcal{G}})  + \lambda_3 \mathcal{F}(\boldsymbol{\mathcal{G}}) )\label{eq:exponential}
\end{equation}

The description of the distribution reflects the characteristics that we would like our potential target interactions to possess. First, the distribution favours interactions with small weights, as larger weights signal interactions that are more frequent with respect to a given object (e.g. patient), and those are less likely to have caused the outcome, as described in the introduction. Hence the exponentially decreasing function of $\mathcal{W}(\boldsymbol{\mathcal{G}})$, which returns in the average weight of a given subgraph. 

Second, the distribution favours larger interaction subgraphs, which is reflected by the exponentially increasing function of $\mathcal{N}(\boldsymbol{\mathcal{G}})$. This may seem counter intuitive, going against the observations concerning the medical domain, where the number of variables leading to an outcome is orders of magnitude less than the number of variables measured. However, we will make the reasoning behind this clear after examining the third component of the distribution. 

Finally, the distribution favours subgraphs with maximum favouritism with respect to the population, which is calculated by aggregating its similarity to other graphs in the population: 

\begin{equation}
    \mathcal{F}({G}) = \log(\prod_{{G}_z \in \boldsymbol{\mathcal{G}} } \mathcal{S} ({G},{G}_z))
\end{equation} 

Hence the final component describing the distribution is an exponentially increasing function of $\mathcal{F}(\boldsymbol{\mathcal{G}})$.

Although the second component $\mathcal{N}$ favours larger subgraphs, its effect is smoothed by the third component $\mathcal{F}$, because as subgraphs become larger, the likelihood of finding similar subgraphs in the population decreases. In fact, the two competing functions $\mathcal{N}$ and $\mathcal{F}$ force the distribution to generate smaller subgraphs to achieve a high favouritism score. 

 The parameters of the independent components of the distribution are used to find the optimal significant interactions of a (latent) outcome for a given dataset, as described in \ref{sec:algorithm} below. 
 
\subsection{Parameter Estimation}\label{sec:algorithm}

Since enumerating all possible interactions and evaluating their significance is infeasible, the EM algorithm \cite{em} is applied to estimate the parameters of the distribution and use those to find significant interpretations of the model. An interpretation $\mathcal{I}$ is a subgraph containing a subset of the interactions in a population of qualitative interaction graphs $\boldsymbol{\mathcal{G}}$. 
There are three parameters in the model: $\lambda_1 - \lambda_3$. Along with the three generative functions $f_{\lambda_1} - f_{\lambda_3}$, they define the distribution generating significant interactions as per Equation \ref{eq:exponential}. The expected value of the log likelihood function $\mathcal{Q}(\lambda,\lambda^*)$ becomes:

\begin{equation*}
    \mathcal{Q}(\lambda,\lambda^*) = \displaystyle \sum_{k=1}^{m} \sum_{i=1}^{3} E_{\lambda_i^*} (\log f_{\lambda_i} \mathcal(I) | G_k)
\end{equation*}

$m$ is the total number of generated graphs, which correspond to the number of objects in the dataset (see Section \ref{sec:qig}). Since each object is described by many records (with each record corresponding to a time-stamped value for a given variable), $m \ll \mathcal{M}$, where $\mathcal{M}$ is the number of records in the raw dataset. Another note is that the functions $\mathcal{N}$ and $\mathcal{W}$ used to partially compute $\mathcal{Q}$ are linear with respect to a given interpretation $\mathcal{I}$. 

For the M-step, we estimate the values of the parameters $\lambda_1 - \lambda_3$ and functions $f_{\lambda_1} - f_{\lambda_3}$ that maximize $\mathcal{Q}(\lambda)$: 
\begin{equation*}
     \lambda^{max} = \argmax \mathcal{Q}(\lambda,\lambda^*) 
\end{equation*}

Upon convergence of the EM procedure, we sample the distribution of interpretations to obtain the significant interactions of the model, giving preference to interpretations interpretations with lower mean edge weight. 

Because iterating over the set of all possible interpretations via the EM algorithm is also infeasible, we include a pre-processing step to obtain clusters of interpretations that are likely to be optimal, using the graph lexicon to cluster the interaction graphs embedding the interactions within the data using self-tuning spectral clustering \cite{clustering}. 

\section{Experimental Evaluation}\label{sec:experiments}

Our experiments are based on the discovery and evaluation of interactions that can best explain a given outcome. To achieve this, we organise our evaluation into two steps: 

\begin{enumerate}
\item Pattern Discovery: by using the model presented here to discover significant patterns describing a population of subjects with a given outcome, comparing the generated patterns with those generated by existing temporal pattern discovery models.  
\item Outcome Discovery: by using the discovered patterns as predictors of a given outcome in a classification experiment, and comparing the performance against those of a number of classification machine learning algorithms whose performance is established for the given domain problem.

\end{enumerate}

\subsection{The Medical Problem}
Sepsis, defined by a life-threatening response to infection and potentially leading to multiple organ failure, is a devastating condition and one of the most significant causes of worldwide morbidity and mortality \cite{sepsis1}. Sepsis is implicated in 6 million deaths annually, with costs totaling \$24 billion in the USA alone \cite{sepsis2}.

Early identification of sepsis is a known crucial factor in improving its outcomes \cite{sepsis3}. Yet, existing sepsis prediction models using machine learning have shown mixed results reflecting the difficulty in pinpointing the factors leading to sepsis onest \cite{sepsis4}, and heterogeneity in populations \cite{sepsis5} and methodologies \cite{sepsis6}.

Recently \cite{shaharsepsis} proposed a temporal data mining approach using gradient and state abstractions of temporal intervals to find the most frequent patterns in a population of septic patients. Comparing those to temporal patterns in  non-septic patients found significant statistical differences. The work does not evaluate the ability of the discovered patterns to perform sepsis prediction; a task which we perform as part of our evaluation. 

\subsection{The Dataset}
We used the Medical Information Mart for Intensive Care III (MIMIC-III) \cite{mimic}, a large and freely-accessible de-identified intensive care database from Boston, Massachusetts, USA. MIMIC-III contains demographics, vital sign measurements, laboratory test results, procedures, medications, caregiver notes and imaging reports recorded over time, in addition to mortality data (both in and out of the hospital) for over 46,520 critical care patients.

We processed the data contained within MIMIC-III to exclude: 1. non-adults, less than 15 years of age at the time of admission, 2. invalid admissions, which frequently correspond to clerical errors and are characterised by the absence of heart rate, incomplete administrative recordings or admission and discharge and no charted observations, and 3. stays shorter than 4 hours, as those tend to have high rates of incomplete data and are therefore of little value for our purpose. We then queried the MIMIC-III database for patients satisfying the conditions the third international definition of sepsis \cite{singer}, under the guidance of two domain experts, and using surrogates of an organ dysfunction component of acute increase in SOFA score beyond 2 points and persistent hypotension (mean blood pressure $<$ 65) requiring vasopressors to maintain mean arterial pressure. 

The resulting set comprises 4,720 ICU records for 4,403 patients (a single patient may have multiple ICU visits). We divided the records into two sets of 2,360 ICU records each. The first set will be used for pattern discovery, while the second will be used for evaluation. In addition, we extracted an equivalent 2,360 records from the records that did not satisfy our inclusion criteria to serve as negative examples for the classification experiments of our evaluation. We extracted the variables shown in Table \ref{tab:features} and used the state and gradient abstraction cutoffs shown in the table to abstract the raw data into more meaningful concepts. The knowledge base given in the table is an exact replica of that used in previous work using temporal knowledge-based temporal abstraction to prediction sepsis \cite{shaharsepsis}.

\begin{table}
\begin{center}
{\caption{The knowledge base detailing the variables and state and gradient cutoff values for temporal abstraction of ICU vitals and laboratory tests}\label{tab:features}}
\begin{tabular}{llr}
\hline
\rule{0pt}{12pt}
\textbf{Clinical} & \textbf{'Normal' State}  & \textbf{Gradient }\\
\textbf{Concept} & \textbf{Abstraction} & \textbf{Abstraction} \\\hline
Albumin & 3.4-5.4 g/dL    & $\Delta$ $>$ 0.5 \\
Bilirubin 	 &0.2 - 1.2 mg/dL  	 & $\Delta$ $>$ 0.5 \\
chloride 	 &96 – 106 mEq/L  	&$\Delta>$5 \\
Fibrinogen 	 &200 – 400 mg/dL  	  &$\Delta$ $>$ 50 \\
Creatinine 	 &0.6-1.3 mg/dL  	  &$\Delta$ $>$ 0.2 \\
Glucose  	  &70 – 100 mg/dL  	  &$\Delta$ $>$ 10 \\
Hemoglobin  	  &11 – 18 g/dL  	  &$\Delta >$2 \\
Lactate 	 &0.5 - 2.2 mmol/L  	  &$\Delta >$1 \\
PCO2  	  &38 – 42 mm Hg  	  &$\Delta >$2 \\
Urea 	 &10 – 20 mg/dL  	  &$\Delta>$5 \\
Sodium 	 &135 – 145 mEq/L  	  &$\Delta>$5 \\
TCO2 	 &22 – 28 mmol/l  	  &$\Delta>$2 \\
WBC  	  &4.5 – 10 $\times$ 10$^9$/L  	  &$\Delta>$1 \\
Body Temperature  	  &36 – 38 $\degree$ C  	  &$\Delta>$0.5 \\
Glasgow Coma Scale 	  & 8-12	 &$\Delta>$2 \\
Diastolic Blood Pressure  	  &70 – 90 mmHg  	  &$\Delta>$10 \\
Systolic Blood Pressure  	  &110 – 140 mmHg  	  &$\Delta>$10 \\
Mean Blood Pressure  	  &65 - 80 	 &$\Delta>$5 \\
Heart Rate 	 &60 – 80 bpm 	 &$\Delta>$10 \\
Spontaneous Respiratory Rate 	 &7 – 14 breath/pm  	  &$\Delta>$3  \\
Platelets 	 &150 – 400 $\times$ 10$^9$/L  	  &$\Delta>$ 50 \\
PO2 (PaO2 in Andrea) 	 &75 – 100 tor  	  &$\Delta>$10 \\
PCO2 (PaCO2 in Andrea) 	 &38 – 42 mm Hg  	  &$\Delta>$2 \\
\hline
\end{tabular}
\end{center}
\end{table}

\subsection{Experiments and Results}
\subsubsection{Discovering Significant Sepsis Patterns}
This experiment aims to discover significant patterns of interactions within the septic population 8 hours prior to sepsis onset (similarly to \cite{shaharsepsis}). We re-queried the dataset to only include vitals which have been recorded over 8 hours before the confirmation of sepsis was recorded in the patient's records. We used the 2,360 records allocated to pattern discovery, which were all septic patients. After filtration, 1,679 records satisfied the temporal constraint enforced on the records.  Summary of the discovered interactions is given in Table \ref{tab:experiment1}. In the table, the first column describes the sampling rate: the percentage of significant interactions sampled from the distribution of interpretations after the convergence of the EM procedure of Section \ref{sec:algorithm}. The second column shows the number of patterns discovered using the corresponding sampling rate in the first column. The third column describes the percentage of septic patients with the sampled patterns being subsets of their qualitative interaction graphs, while the fourth column describes the same percentage non-septic patients.

The table clearly shows that the top 5\% of the discovered interactions are almost only exclusively found in the septic population. Moreover, comparing these results with the temporal patterns discovered via frequent pattern mining in \cite{shaharsepsis} shows the clear contrast in the two approaches and highlights the advantages of searching for rare interactions. \cite{shaharsepsis} found a total of 6,168 patterns with exclusive prevalence in the septic population and 14,384 patterns found to be present in both septic and non-septic patients.

\begin{table}
\begin{center}
{\caption{The distribution of significant patterns across septic and non-septic patients}\label{tab:experiment1}}
\begin{tabular}{llll}
\hline
\rule{0pt}{12pt}
\textbf{Sampling}  & \textbf{Number of} & \textbf{Prevalence in } & \textbf{Prevalence in} \\
        \textbf{Rate} &  \textbf{Patterns} &  \textbf{Septic Patients} & \textbf{Non-septic Patients} \\\hline
5\%    &  39  & 99\%      & $<$ 0.05\% \\
10\%   &  75 &  94.8\% &  0.08\%\\
40 \% &  302 & 0.82\%& 0.2\%\\
80\% & 610  & 0.79\% & 0.26\%\\
    
\end{tabular}
\end{center}
\end{table}

We note that the visible improvement in the number and discriminatory power of the discovered interactions is achieved despite the fact that our EM procedure is unsupervised, and was only given a population of septic patients (no negative examples or labeling performed). This is in contrast to \cite{shaharsepsis}, which performed the pattern discovery process using clearly labeled sepsis and non-sepsis records. 

Given the stark difference in discriminatory power of discovered interactions, we argue that the QIG model not only enables the discovery of more useful interactions, but the semantics of the model captured by the joint representation of qualitative state and gradient change highly contribute to the derivation of a smaller and more meaningful set of significant interactions.

\begin{table}
\begin{center}
{\caption{Prevalence of interaction subsets in the QIG most significant patterns}\label{tab:prevalence}}
\begin{tabular}{llll}
\hline
\rule{0pt}{12pt}
\textbf{Node 1} & \textbf{Node 2} & \textbf{Temporal Rel} & \textbf{Prevalence} \\\hline
 SysBP-Low-Dec & Temp-Hi-Incr & Finishes & 77\% \\ 
  WBC-Norm-Inc & Temp-Hi-Incr & Overlaps & 61\% \\ 
  Heartrate-Hi-Stab& Temp-Hi-Stab & Overlaps & 49\% \\ 
Lactate-Hi-Stab & Heartrate-Hi-Stab & During & 35\% \\ 
 
\end{tabular}
\end{center}
\end{table}

We further examined the top 5\% (39) patterns discovered by our QIG model for the clinical significance of the most prevalent subsets of interactions. These are given in Table \ref{tab:prevalence}.

\begin{table*}[ht!]
\begin{center}
{\caption{Comparison of performance metrics of the QIG model in predicting sepsis 8 hours prior to ICU admission, with the performance of state of the art sepsis prediction models obtained from the literature. AUROC: Area Under the Receiver-operator Curve. PLR: Positive Likelihood Ratio ($\displaystyle\frac{sensitivity}{1-specificity}$), NLR: Negative Likelihood Ratio ($\displaystyle\frac{1-sensitivity}{specificity}$). Note: * indicates values calculated using the reported sensitivity and specificity.}\label{tab:experiment2}}
\begin{tabular}{llllll}
\hline
\rule{0pt}{12pt}
\textbf{Model}   & \textbf{Sensitivity} & \textbf{Specificity} & \textbf{AUROCC} & \textbf{PLR} & \textbf{NLR} \\
        \textbf{} &   \textbf{(95\% CI)} & \textbf{ 95\% CI} & &\textbf{(95\% CI)} & \textbf{(95\% CI)}  \\\hline
        
QIG & 0.98 (0.96-0.99)  & 0.95 (0.91 - 0.99) & 0.97 & 19.6 & 0.02 \\
Kam & 0.94 (0.93-0.95) & 0.91 (0.88-0.94) & 0.929 & 10.45$^*$ & 0.07$^*$ \\
Mao & 0.98 (0.96- 0.99) & 0.8 (0.78-0.82) & 0.915 & 4.90$^*$ & 0.03$^*$ \\
Desautels & 0.80 (0.79-0.81)& 0.79 (0.78–0.81) & 0.791 &3.8$^*$  & 0.25$^*$  \\
Nemati & 0.85 (0.84 - 0.86) & 0.84 (0.82 - 0.86) &  0.85 & 5.3$^*$  & 0.18$^*$  \\

Calvert & 0.9 (0.89–0.91) & 0.81 (0.80–0.82) &  0.86 & 4.7$^*$  & 0.12$^*$  \\\hline

\end{tabular}
\end{center}
\end{table*}

Close examination of the most prevalent interaction subsets (with the help of two domain experts) show that the variables and the order of their changes in the interactions is semantically meaningful from a clinical perspective. For instance, the first pattern describes periods where a patient has a low and consistently decreasing systolic blood pressure, such that these periods are temporally equivalent (in a qualitative sense) to the finishing intervals of increasing high temperature. Clinically, having a high temperature is one of the first alarming signs in an ICU patient and is a valid reason to look for signs of infections. As the possibility of an infection increases in a given patient, her other vitals start showing signs of deterioration, and a low and decreasing Systolic blood pressure is one of the first to show significant values changes. Similarly, increasing (yet still in the normal range) white blood cell count indicate that the body is preparing to fight an infection. However, if fever develops during this period of incremental increase in WBC count, this may indicate that the body is in alarm mode and an infection maybe imminent.

\subsubsection{Using Significant Patterns as Predictors of Sepsis}

Using the set of 2,360 ICU records of septic patients not used in pattern discovery as positive examples, and the 2,360 ICU records of non-septic patients as negative examples, and further filtering the data to exclude variables recorded less than 8 hours of confirmed sepsis diagnosis, we performed a classification task with sepsis diagnosis as outcome and using the top 10\% interaction patterns as features for the learning task (for a total of 39 features). Using the XGBoost classification algorithm,  the classifier's parameters were optimised through a bootstrapped grid search over is hyperparameter space. After filtering the records by time, our dataset contained a substantial class imbalance (only 36\% positive examples). We therefore employed cost-sensitive learning by placing a heavier penalty on misclassifying the minority class (septic patients). The classifier was training over 1,000 iterations of a 10-fold cross-validation, incorporating class weights into each. 

 In addition to the performance metrics of the QIG-based classifier reported in Table \ref{tab:experiment2}, we collected the reported performance metrics of five models representative of current sepsis predictors: \cite{kam} (referred to as Kam), \cite{mao} (referred to as Mao), \cite{Desautels} (referred to as Desautels), \cite{Nemati} (referred to as Nemati), and \cite{calvert} (referred to as Calvert). Indeed, the results underline the current bottleneck of ML-based sepsis prediction: the highest performance of the models available in the literature was reported by Mao and Kam. Mao relies on feature selection to achieve performance, but only reports high sensitivity at the fixed specificity value of 0.8. In contrast, Kam is a neural network model which does not perform explicit feature selection. While it reports high sensitivity and specificity, the black-box nature of the Kam model hinders its clinical utility in practical settings. It is worth noting that the improved performance of the QIG model is especially pronounced in the specificity of predictions, in which more explainable models (Mao) underperforms by not distinguishing septic patients from those with inflammations and comorbidities; a general bottleneck in ML sepsis prediction \cite{islam}. It is also worth noting that only \cite{Nemati} claims to be a fully interpretable model (by virtue of feature importance). 

\section{Conclusions, Limitations and Future Work} \label{sec:conclusion}

In this work, we have proposed a framework for the discovery of rare patterns of interactions from highly-dimensional, multivariate and non-uniformly sampled time series data. The model uses qualitative abstraction to capture meaningful semantics embedded within the raw data and formulates a probabilistic model governing rare interactions. We used an expectation-maximisation procedure to uncover temporal interactions with the highest contributions to a given outcome. 
The paper presents experiments using the framework on real ICU data to discover rare temporal interactions contributing to the onset of sepsis. The results show that using the patterns identified by our model as features in classification experiments yields a superior discriminatory power to when using the raw data, as well as superior performance to state of the art machine learning algorithms that have been specifically optimised for sepsis prediction. 

Our present research focuses on a number of areas of improvement. To begin with, QIGs learned from the data can be further optimised using the topological constraints of Allen's relations to remove redundant labels by inferring them from the graph \cite{redundancy}. Moreover, using a multi-resolutional representation of time to create QIGs can accommodate imprecise, gradual and intuitive relations between points and intervals \cite{granular}.  We are also designing a causal rare event discovery framework exploiting the rich semantics captured by our representation. Finally, we are working on the algorithmic aspects of pattern discovery to devise efficient and scalable algorithms for the fast detection of rare temporal interactions, overcoming the slow EM algorithm implementation, which is far from the ultimate goal of real-time monitoring.

\section{Acknoweldgements}
We sincerely thank Andrea Agarossi and Ahmed Hamoud for their valuable insight and domain expertise. This research was supported by the following funding bodies: 

\begin{enumerate}

\item  Health Data Research UK, which is funded by the UK Medical Research Council, Engineering and Physical Sciences Research Council, Economic and Social Research Council, Department of Health and Social Care (England), Chief Scientist Office of the Scottish Government Health and Social Care Directorates, Health and Social Care Research and Development Division (Welsh Government), Public Health Agency (Northern Ireland), British Heart Foundation and Wellcome Trust.
\item The National Institute for Health Research University College London Hospitals Biomedical Research Centre.
\item  National Institute for Health Research (NIHR) Biomedical Research Centre at South London and Maudsley NHS Foundation Trust and King’s College London. 

\item The UK Medical Research Council (MRC), grant numbers MR/S004149/1 and MC\_PC\_18029.

\end{enumerate}

\bibliography{ecai}

\begin{thebibliography}{10}

\bibitem{allen1983}
James Allen, `Maintaining knowledge about temporal intervals', {\em
  Communications of the {ACM}}, {\bf 26}(11),  832--843, (1983).

\bibitem{calvert}
Jacob Calvert, Daniel Price, Uli Chettipally, and et~al, `A computational
  approach to early sepsis detection', {\em Comput. Biol. Med.}, {\bf 74}(C),
  69–73, (July 2016).

\bibitem{granular}
Quentin Cohen-Solal, Maroua Bouzid, and Alexandre Niveau, `An algebra of
  granular temporal relations for qualitative reasoning', in {\em Proceedings
  of the Twenty-Fourth International Joint Conference on Artificial
  Intelligence {(IJCAI)}}, pp. 2869--2875, (2015).

\bibitem{Desautels}
Thomas Desautels, Jacob Calvert, Jana Hoffman, and et~al r, `Prediction of
  sepsis in the intensive care unit with minimal electronic health record data:
  A machine learning approach', {\em {JMIR} Medical Informatics}, {\bf 4}(3),
  e28, (Sep 2016).

\bibitem{sepsis6}
{Norman Lance} Downing, Joshua Polnick, Sarah Poole, and et~al, `Electronic
  health record-based clinical decision support alert for severe sepsis', {\em
  {BMJ} Quality Improvement}, {\bf 1},  1--7, (2019).

\bibitem{sepsis1}
Carolin Fleischmann, André Scherag, Neill~Kj Adhikari, and et~al, `Assessment
  of global incidence and mortality of hospital-treated sepsis. current
  estimates and limitations', {\em American Journal of Respiratory and Critical
  Care Medicine}, {\bf 193}(3),  259--–72, (2016).

\bibitem{medicalqualitative5}
John Fox, David Glasspool, and Jonathan Bury, `Quantitative and qualitative
  approaches to reasoning under uncertainty in medical decision making', in
  {\em Proceedings of the 8th Conference on AI in Medicine in Europe:
  Artificial Intelligence Medicine}, AIME '01, pp. 272--282, (2001).

\bibitem{freska}
Christian Freksa, `Conceptual neighborhood and its role in temporal and spatial
  reasoning', in {\em Proceedings of the IMACS Workshop on Decision Support
  Systems and Qualitative Reasoning}, pp. 181--187, (1991).

\bibitem{sepsis2}
David Gaieski, Matthew Edwards, Michael Kallan, and Brendan Carr, `Benchmarking
  the incidence and mortality of severe sepsis in the united states', {\em
  Critical Care Medicine}, {\bf 41}(5),  1167--–1174, (2013).

\bibitem{allenalgebra}
Michael Gr{\"u}ninger and Zhuojun Li, `The time ontology of allen's interval
  algebra', in {\em TIME}, (2017).

\bibitem{em}
Maya Gupta and Yihua Chen, `Theory and use of the em algorithm', {\em
  Foundations and Trends in Signal Processing}, {\bf 4}(3),  223–--296,
  (2010).

\bibitem{hu2013}
Bing Hu, Yanping Chen, and Eamonn Keogh, `Time series classification under more
  realistic assumptions', in {\em Proceedings of SIAM Data Mining}, (2013).

\bibitem{ibrahim}
Zina Ibrahim, Ahmed Tawfik, and Alioune Ngom, `Qualitative motif detection in
  gene regulatory networks', in {\em 2009 IEEE International Conference on
  Bioinformatics and Biomedicine}, pp. 124--129, (2009).

\bibitem{medicalqualitative3}
Liliana Ironi, Mario Stefanelli, and Giordano Lanzola, `Qualitative models in
  medical diagnosis', {\em Artificial Intelligence in Medicine}, {\bf 2}(2),
  85--101, (1990).

\bibitem{islam}
{Mohammed Mohaimenul} Islam, Tahmina Nasrin, {Bruno Andreas} Walther, and
  et~al, `Prediction of sepsis patients using machine learning approach: A
  meta-analysis', {\em Computer Methods and Programs in Biomedicine}, {\bf
  170},  1--9, (2019).

\bibitem{mimic}
Alistair Johnson, Tom Pollard, Lu~Shen, and et~al, `Mimic-iii, a freely
  accessible critical care database', {\em Scientific Data}, {\bf 3},  160035,
  (2016).

\bibitem{kam}
{Hye Jin } Kam and { Ha Young} Kim, `Learning representations for the early
  detection of sepsis with deep neural networks.', {\em Computational Methods
  for Biology and Medicine.}, {\bf 89}(1),  248--–55, (2017).

\bibitem{kleinberg2011}
Samantha Kleinberg, `A logic for causal inference in time series with discrete
  and continuous variables', in {\em In Proceedings of the 22nd International
  Joint Conference on Artificial Intelligence {(IJCAI)}}, pp. 943--950, (2011).

\bibitem{kleinberg2012}
Samantha Kleinberg, {\em Causality, Probability, and Time}, Cambridge
  University Press, 2012.

\bibitem{causal}
Samantha Kleinberg, `Causal inference with rare events in large-scale
  time-series data', in {\em Proceedings of the Twenty-Third International
  Joint Conference on Artificial Intelligence {(IJCAI)}}, pp. 1444--1450,
  (2013).

\bibitem{kleinberg2009}
Samantha Kleinberg and Bud Mishra, `The temporal logic of causal structures',
  in {\em In Proceedings of the 25th Conference on Uncertainty in Artificial
  Intelligence}, pp. 303--312, (2009).

\bibitem{medicalqualitative4}
Benjamin Kuipers, `New reasoning methods for artificial intelligence in
  medicine', {\em International Journal of Man-Machine Studies}, {\bf 26}(6),
  707--718, (1987).

\bibitem{redundancy}
Sanjiang Liab, Zhiguo Long, Weimin Li, Matt Duckham, and Alan Both, `On
  redundant topological constraints', {\em Artificial Intelligence}, {\bf 225},
   51 -- 76, (2015).

\bibitem{mao}
Qingqing Mao, Melissa Jay, Jana Hoffman, and et~al, `Validation of a sepsis
  prediction algorithm using only vital sign data in the emergency department,
  general ward and {ICU}', {\em BMJ Open}, {\bf 8}(1),  e017833, (2018).

\bibitem{sepsis5}
Andrea McCoy and Ritankar Das, `Reducing patient mortality, length of stay and
  readmissions through machine learning-based sepsis prediction in the
  emergency department, intensive care unit and hospital floor units', {\em
  {BMJ} Open Quality}, {\bf 6}(2),  e000158, (2017).

\bibitem{moskovitch2015}
Robert Moskovitch and Yuval Shahar, `Classification of multivariate time series
  via temporal abstraction and time intervals mining', {\em Knowledge and
  Information Systems}, {\bf 45}(1),  35--74, (Oct 2015).

\bibitem{shahar2015}
Robert Moskovitch and Yuval Shahar, `Fast time intervals mining using the
  transitivity of temporal relations', {\em Knowledge Information Systems},
  {\bf 42}(1),  21--48, (2015).

\bibitem{murphy2002}
Kevin Murphy, {\em Dynamic Bayesian networks: representation, inference and
  learning}, University of California, Berkley, 2002.

\bibitem{Nemati}
Shamim Nemati, Andre Holder, Fereshteh Razmi, and et~al, `An interpretable
  machine learning model for accurate prediction of sepsis in the {ICU}', {\em
  Critical care medicine}, {\bf 46}(4),  547--553, (April 2018).

\bibitem{medicalqualitative2}
Kristina Orban, Maria Ekelin, Gudrun Edgren, and et~al, `Monitoring progression
  of clinical reasoning skills during health sciences education using the case
  method – a qualitative observational study', {\em {BMC} Medical Education},
  {\bf 17}(158),  e27, (2017).

\bibitem{patel2008}
Dhaval Patel, Wynne Hsu, and Mong-Li Lee, `Mining relationships among
  interval-based events for classification', in {\em Proceedings of the {ACM
  SIGMOD} international conference on Management of data}, pp. 393--–404,
  (2008).

\bibitem{pearl}
Judea Pearl, {\em Causality: Models, Reasoning, and Inference}, Cambridge
  University Press, 2000.

\bibitem{medicalqualitative1}
Clara Schaarup and {Louise Bilenberg} Pape-Haugaard and{ Ole Kristian}
  Hejlesen, `Models used in clinical decision support systems supporting
  healthcare professionals treating chronic wounds: Systematic literature
  review', {\em Medical Informatics Research Diabetes}, {\bf 3}(2),  e11,
  (2018).

\bibitem{sepsis3}
Christopher Seymour, Foster Gesten, Hallie Prescott, and et~al, `Time to
  treatment and mortality during mandated emergency care for sepsis', {\em The
  New England Journal of Medicine}, {\bf 376}(23),  2235–--2244, (2017).

\bibitem{shaharkb}
Yuval Shahar, `A framework for knowledge-based temporal abstraction', {\em
  Artificial Intelligence}, {\bf 90}(1–2),  79--133, (1997).

\bibitem{shaharsepsis}
Eitam Sheetrit, Nir Nissim, Denis Klimov, and et~al.
\newblock Temporal pattern discovery for accurate sepsis diagnosis in icu
  patients.
\newblock arXiv.org.

\bibitem{shahar2017}
Alexander Shknevsky, Yuval Shahar, and Robert Moskovitch, `Consistent discovery
  of frequent interval-based temporal patterns in chronic patients’ data',
  {\em Journal of Biomedical Informatics}, {\bf 75},  83--95, (2017).

\bibitem{singer}
Mervyn Singer, Clifford Deutschman, Christopher~Warren Seymour, and et~al, `The
  third international consensus definitions for sepsis and septic shock
  (sepsis-3)', {\em {JAMA}}, {\bf 315}(8),  801--–810, (2016).

\bibitem{aime}
Marion Verduijn, Arianna Dagliati, Lucia Sacchi, and et~al, `Comparison of two
  temporal abstraction procedures: a case study in prediction from monitoring
  data.', in {\em AMIA - Annual Symposium proceedings / AMIA Symposium. AMIA
  Symposium}, pp. 749--753, (2005).

\bibitem{sepsis4}
Jean-Louis Vincent, `The clinical challenge of sepsis identification and
  monitoring', {\em {PLoS} Medicine}, {\bf 13}(5),  e1002022, (2016).

\bibitem{winarko}
Edi Winarko and John Roddick, `Armada – an algorithm for discovering richer
  relative temporal association rules from interval-based data', {\em Data and
  Knowledge Engineering}, {\bf 63}(1),  76--90, (2007).

\bibitem{clustering}
Lihi Zelnik-Manor and Pietro Perona, {\em Self-tuning spectral clustering},
  1601--–1608, Advances in Neural Information Processing Systems, 2005.

\end{thebibliography}
\end{document}